\documentclass{article} 
\usepackage{iclr2016_conference,times}
\usepackage{hyperref}
\usepackage{url}
\usepackage{amsmath}
\usepackage{color}
\usepackage{url}
\usepackage{multirow}
\usepackage{symbols}
\usepackage{framed}
\usepackage{graphicx}

\usepackage{algorithm}
\usepackage{algorithmicx}
\usepackage{algpseudocode}
\usepackage{wrapfig}
\usepackage{alltt}
\usepackage{booktabs}
\usepackage[activate={true,nocompatibility},final,kerning=true,spacing=true,factor=1100,stretch=10,shrink=10]{microtype}

\usepackage{tikz}
\usetikzlibrary{arrows,decorations.pathmorphing,backgrounds,fit,automata,shapes,calc,positioning,shadows}
\tikzset{heapgraph/.style={
                every node/.style={rectangle, font=\footnotesize},
                tree/.append style={draw=black, thick, minimum width=.4cm, fill=blue!20},
                list/.append style={
                  draw=black,
                  thick,
                  minimum width=.4cm,
                  fill=green!20, 
                  prefix after command= {\pgfextra{\tikzset{every label/.style={font=\small, inner sep=2pt}}}}},
                active/.append style={double},
                explained/.append style={fill=white},
                colLabel/.append style={font=\small},
                stepLabel/.append style={text width=2cm,align=left,font=\small},
                outLabel/.append style={font=\small},
                heapEdgeLabel/.append style={font=\tiny,inner sep=2pt},
        },
}

\newcommand{\heapGNode}[0]{\ensuremath{v}}

\newcommand{\SLls}[0]{\ensuremath{\mathsf{ls}}}
\newcommand{\SLtree}[0]{\ensuremath{\mathsf{tree}}}
\newcommand{\SLNone}[0]{\ensuremath{\top}}
\newcommand{\SLempty}[0]{\ensuremath{\mathsf{none}}}
\newcommand{\SLvar}{\ensuremath{\mathsf{var}}}
\newcommand{\SLnull}{\texttt{NULL}}
\newcommand{\SLcurr}{\texttt{curr}}
\newcommand{\SLlst}{\texttt{lst}}
\newcommand{\SLelt}{\texttt{elt}}

\newcommand{\OurMethodMinor}{Gated Graph Neural Network}
\newcommand{\OurMethodMinors}{\OurMethodMinor s}
\newcommand{\OurMethodMinorShort}{GG-NN}
\newcommand{\OurMethodMinorShorts}{\OurMethodMinorShort s}
\newcommand{\OurMethod}{Gated Graph Sequence Neural Network}
\newcommand{\OurMethods}{\OurMethod s}
\newcommand{\OurMethodShort}{GGS-NN}
\newcommand{\OurMethodShorts}{\OurMethodShort s}

\newcommand\algoutput[0]{\textbf{print}}

\title{\OurMethods}

\author{Yujia Li\thanks{Work done primarily while author was an intern at
    Microsoft Research.}\ \ \&
Richard Zemel
\\
Department of Computer Science, University of Toronto \\
Toronto, Canada \\
\texttt{\{yujiali,zemel\}@cs.toronto.edu} \\
\AND
Marc Brockschmidt \& Daniel Tarlow \\
Microsoft Research \\
Cambridge, UK \\
\texttt{\{mabrocks,dtarlow\}@microsoft.com}
}

\iclrfinalcopy 

\begin{document}


\maketitle

\begin{abstract}
  Graph-structured data appears frequently in
  domains including chemistry, natural language semantics, social
  networks, and knowledge bases. 
  In this work, we study
  feature learning techniques for graph-structured inputs. Our
  starting point is previous work on Graph Neural Networks
  \citep{scarselli2009graph}, which we modify to use gated recurrent
  units and modern optimization techniques and then extend
  to output sequences. The result is a flexible and broadly
  useful class of neural network models that has favorable inductive
  biases relative to purely sequence-based models (e.g.,~LSTMs) when the
  problem is graph-structured.  We demonstrate the capabilities on some
  simple AI (bAbI) and graph algorithm learning tasks. We then show it
  achieves state-of-the-art performance on a problem from program verification,
  in which subgraphs need to be described as abstract data structures.
\end{abstract}

\comment{
\begin{abstract}
  Much data in the real world is naturally represented as a graph.  For
example, entities can be represented as nodes, with relations between
them represented as edges. Graph-structured data comes up often in
chemistry, social networks, knowledge bases, and web-based data, and
in other domains like program verification which requires reasoning
about abstract data structures represented as graphs.  In this work,
we study feature learning techniques for learning from
graph-structured inputs. Our starting point is previous work on Graph
Neural Networks (GNNs) \citep{scarselli2009graph}. We begin by adapting the model and
learning algorithm to make use of recent advances in recurrent neural
networks, incorporating gated recurrent units and modern optimization
routines. Our main contribution is then to extend GNNs to output
sequences. We use bAbI tasks to illustrate the components of the model
and explain what is being learned by the GNN models, and then we
develop a full model that learns to predict loop invariants for a
program verification task, where we map from graphs representing the
state of memory during a program's execution to a logical description
of the data structures that have been instantiated. We show that we
are able to match the performance of a system that has been heavily
hand-engineered, indicating the \OurMethods~are promising models for
feature learning on graphs.
\end{abstract}
}


\section{Introduction}

Many practical applications build on graph-structured data, and thus we often
want to perform machine learning tasks that take graphs as inputs. Standard approaches
to the problem include engineering custom features of an input graph,
graph kernels
\citep{kashima2003marginalized,shervashidze2011weisfeiler}, and
methods that define graph features in terms of random walks on graphs
\citep{perozzi2014deepwalk}.  More closely related to our goal in this
work are methods that learn features on graphs, including Graph Neural
Networks \citep{gori2005new,scarselli2009graph}, spectral networks
\citep{bruna2013spectral} and recent work on learning graph
fingerprints for classification tasks on graph representations of
chemical molecules \citep{duvenaud2015convolutional}.


Our main contribution is an extension of Graph Neural Networks that outputs
sequences. Previous work
on feature learning for graph-structured inputs has focused on models
that produce single outputs such as graph-level classifications, but
many problems with graph inputs require outputting sequences.
Examples include paths on a graph, enumerations of graph nodes with
desirable properties, or sequences of global classifications mixed
with, for example, a start and end node.  We are not aware of existing
graph feature learning work suitable for this problem. Our motivating
application comes from program verification and requires outputting
logical formulas, which we formulate as a sequential output problem.
A secondary contribution is highlighting that
Graph Neural Networks (and further extensions we develop here) are a
broadly useful class of neural network model that is applicable to many problems
currently facing the field.

There are two settings for feature learning on graphs: (1) learning a
representation of the input graph, and (2) learning representations of
the internal state during the process of producing a sequence of outputs.
Here, (1) is mostly achieved by previous work on Graph Neural Networks
\citep{scarselli2009graph}; we make several minor adaptations of
this framework, including changing it to use modern practices around
Recurrent Neural Networks. (2) is important because we desire outputs
from graph-structured problems that are not solely individual
classifications.  In these cases, the challenge is how to learn
features on the graph that encode the partial output sequence that has
already been produced (e.g., the path so far if outputting a path) and
that still needs to be produced (e.g., the remaining path). We will
show how the GNN framework can be adapted to these settings, leading
to a novel graph-based neural network model that we call
\OurMethods~(\OurMethodShorts).

We illustrate aspects of this general model in experiments on bAbI tasks
\citep{weston2015towards} and graph algorithm
learning tasks that illustrate the capabilities of the model.
We then present an application to the verification of computer programs.
When attempting to prove properties such as \emph{memory safety} (i.e., that
there are no null pointer dereferences in a program), a core problem 
is to find mathematical descriptions of the data structures used in a program.
Following \cite{brockschmidt2015learning}, we have
phrased this as a machine learning problem where we will learn to map
from a set of input graphs, representing the state of memory, to a
logical description of the data structures that have been
instantiated.  Whereas \cite{brockschmidt2015learning} relied on a large amount of
hand-engineering of features, we show that the system can
be replaced with a \OurMethodShort~at no cost in accuracy.

\section{Graph Neural Networks}

In this section, we review Graph Neural Networks (GNNs)
\citep{gori2005new,scarselli2009graph} and introduce notation and concepts that will be
used throughout.

GNNs are a general neural network architecture defined according to a
graph structure $\graph = (\nodes, \edges)$.  Nodes $\node \in \nodes$
take unique values from $1, \ldots, |\nodes|$, and edges are pairs
$e = (\node, \node') \in \nodes \times \nodes$.  We will focus in this
work on directed graphs, so $(\node, \node')$ represents a directed
edge $\node \rightarrow \node'$, but we note that the framework can
easily be adapted to undirected graphs; see \cite{scarselli2009graph}.  The
\emph{node vector} (or \emph{node representation} or \emph{node
  embedding}) for node $\node$ is denoted by $\noderep{\node} \in
\reals^D$.  Graphs may also contain node labels $\labels_{\node} \in \{1,
\ldots, \numNodeLabels \}$ for each node $\node$ and edge labels or edge types
$\labels_{\edge} \in \{1, \ldots, \numEdgeLabels\}$ for each edge.
We will overload notation and let 
 $\noderep{\mathcal{S}} = \{ \noderep{\node} \given \node \in \mathcal{S} \}$
when $\mathcal{S}$ is a set of nodes, and
$\labels_{\mathcal{S}} = \{ \labels_{\edge} \given \edge \in
\mathcal{S} \}$ when $\mathcal{S}$ is a set of edges.  The function
$\inc(\node) = \{\node' \given (\node', \node) \in \edges\}$ returns the
set of predecessor nodes $\node'$ with $\node' \rightarrow \node$.
Analogously, $\out(\node) = \{\node' \given (\node, \node') \in \edges\}$ is the
set of successor nodes $\node'$ with edges $\node \rightarrow \node'$.
The set of all nodes
neighboring $\node$ is $\neighbor(\node) = \inc(\node) \cup
\out(\node)$, and the set of all edges incoming to or outgoing from
$\node$ is 
$\co(\node) = \{(\node', \node'') \in \edges \given \node = \node' \lor \node = \node'' \}$.

GNNs map graphs to outputs via two steps. First, there is a propagation
step that computes node representations for each node; second, an
output model $o_\node = g(\noderep{\node}, \labels_{\node})$ 
maps from node representations and corresponding labels to an output $o_{\node}$
for each $\node \in \nodes$. In the notation for $g$, we leave the dependence on
parameters implicit, and we will continue to do this throughout.
The system is differentiable from end-to-end,
so all parameters are learned jointly using gradient-based optimization.

\subsection{Propagation Model}
Here, an iterative procedure propagates node representations.
Initial node representations $\noderept{\node}{1}$ are set to arbitrary values,
then each node representation is updated following the recurrence below until
convergence, where $t$ denotes the timestep:
\begin{align*}
    \noderept{\node}{t} & = f^*(\labels_{\node}, 
                      \labels_{\co(\node)},
                      \labels_{\neighbor(\node)},
                      \noderept{\neighbor(\node)}{t-1}
                      ).
\end{align*}
Several variants are discussed in \cite{scarselli2009graph}
including positional graph forms, node-specific updates, and
alternative representations of neighborhoods. Concretely,
\cite{scarselli2009graph} suggest decomposing $f^*(\cdot)$ to be
a sum of per-edge terms:
\begin{align*}
f^*(\labels_{\node}, 
  \labels_{\co(\node)},
  \labels_{\neighbor(\node)},
  \noderept{\neighbor(\node)}{t}) & = \!\!\!\!\!\!
  \sum_{\node' \in \inc(\node)} \!\!\!\!\!
  f(\labels_{\node}, \labels_{(\node', \node)}, 
  \labels_{\node'},\noderept{\node'}{t-1})
  +\!\!\!\!\!\!\!\!\sum_{\node' \in \out(\node)} \!\!\!\!\!\!\!
  f(\labels_{\node}, \labels_{(\node, \node')}, 
  \labels_{\node'},\noderept{\node'}{t-1}),
\end{align*}
where $f(\cdot)$ is either a linear function of $\noderep{\node'}$
or a neural network. The parameters of $f$ depends on the configuration of
labels, e.g. in the following linear case, $\Av$ and $\bv$ are learnable
parameters,
\begin{align*}
  f(\labels_{\node}, \labels_{(\node', \node)}, 
  \labels_{\node'},\noderept{\node'}{t})
  & = \Av^{(\labels_{\node}, \labels_{(\node', \node)}, \labels_{\node'})}
  \noderept{\node'}{t-1}
  + \bv^{(\labels_{\node}, \labels_{(\node', \node)}, \labels_{\node'})}.
\end{align*}

\subsection{Output Model and Learning}
The output model is defined per node and is a differentiable function 
$g(\noderep{\node}, \labels_{\node})$ that maps to an output. This is
generally a linear or neural network mapping.
\cite{scarselli2009graph} focus on
outputs that are independent per node, which are implemented by
mapping the final node representations $\noderept{\node}{T}$, to an
output $o_{\node} = g(\noderept{\node}{T}, \labels_{\node})$ for each node $v
\in \nodes$. To handle graph-level classifications, they suggest to
create a dummy ``super node'' that is connected to all other nodes by
a special type of edge. Thus, graph-level regression or classification
can be handled in the same manner as node-level regression or
classification.

Learning is done via the Almeida-Pineda algorithm
\citep{almeida1990learning,pineda1987generalization}, which works by
running the propagation to convergence, and then computing gradients
based upon the converged solution. This has the
advantage of not needing to store intermediate states in order to
compute gradients. The disadvantage is that 
parameters must be constrained so that
the propagation step is a contraction map. This is needed to ensure
convergence, but it may limit the expressivity of the model.
When $f(\cdot)$ is a neural
network, this is encouraged using a penalty term on the 1-norm
of the network's Jacobian.
  See Appendix~\ref{appendix:contraction-example} for an
example that gives the intuition that contraction maps have trouble
propagating information across a long range in a graph.


\section{\OurMethodMinors}

We now describe \OurMethodMinors~(\OurMethodMinorShorts), our
adaptation of GNNs that is suitable for non-sequential outputs. We
will describe sequential outputs in the next section. The biggest
modification of GNNs is that we use Gated Recurrent Units
\citep{cho2014learning} and unroll the recurrence for a fixed number of steps $T$ and use
backpropagation through time in order to compute gradients. This
requires more memory than the Almeida-Pineda algorithm, but it removes
the need to constrain parameters to ensure convergence. We also
extend the underlying representations and output model.

\subsection{Node Annotations}

In GNNs,
there is no point in initializing node representations because the
contraction map constraint ensures that the fixed point is independent
of the initializations. This is no longer the case with \OurMethodMinorShorts,
which lets us incorporate node labels as additional inputs. To distinguish
these node labels used as inputs from the ones introduced before, we call them
\emph{node annotations}, and use vector $\labelsv$ to denote these
annotations. 


To illustrate how the node annotations are used, consider an example task of
training a graph neural network to predict whether
node $t$ can be reached from node $s$ on a given graph.  For this task,
there are two problem-related special nodes, $s$ and $t$. To mark these nodes
as special, we give them an initial annotation. The first node $s$ gets the
annotation $\labelsv_s=[1,0]^\top$, and the second node $t$ gets the annotation
$\labelsv_t=[0,1]^\top$. All other nodes $\node$ have their initial annotation set to
$\labelsv_\node=[0,0]^\top$. Intuitively, this marks $s$ as the first input argument and $t$ as the second input argument.
We then initialize the node state vectors $\noderept{\node}{1}$ using these label vectors
by copying $\labelsv_\node$ into the first dimensions and padding with extra 0's to allow hidden states that are larger than the annotation size.

In the reachability example, it is easy for the propagation model to learn to propagate the node annotation
for $s$ to all nodes reachable from $s$, for example by setting the propagation matrix associated with forward edges to have a 1 in position (0,0). This will cause the first dimension of node representation to be copied along forward edges.
With this setting of parameters, the propagation step will cause all nodes
reachable from $s$ to have their first bit of node representation set to 1.
The output step classifier can then easily tell whether node $t$ is reachable from $s$ by looking whether some node has nonzero entries in the first two dimensions of its representation vector.


\begin{figure}[t]
\begin{center}
\begin{tabular}{ccc}
\includegraphics[width=.25 \columnwidth]{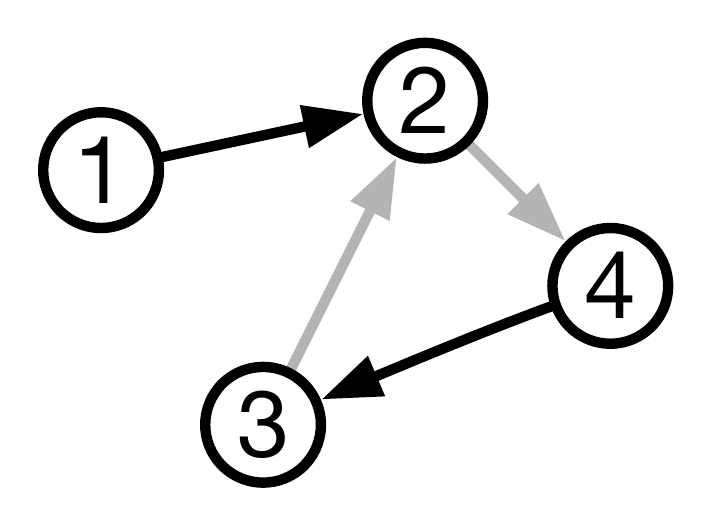} &
\includegraphics[width=.18 \columnwidth]{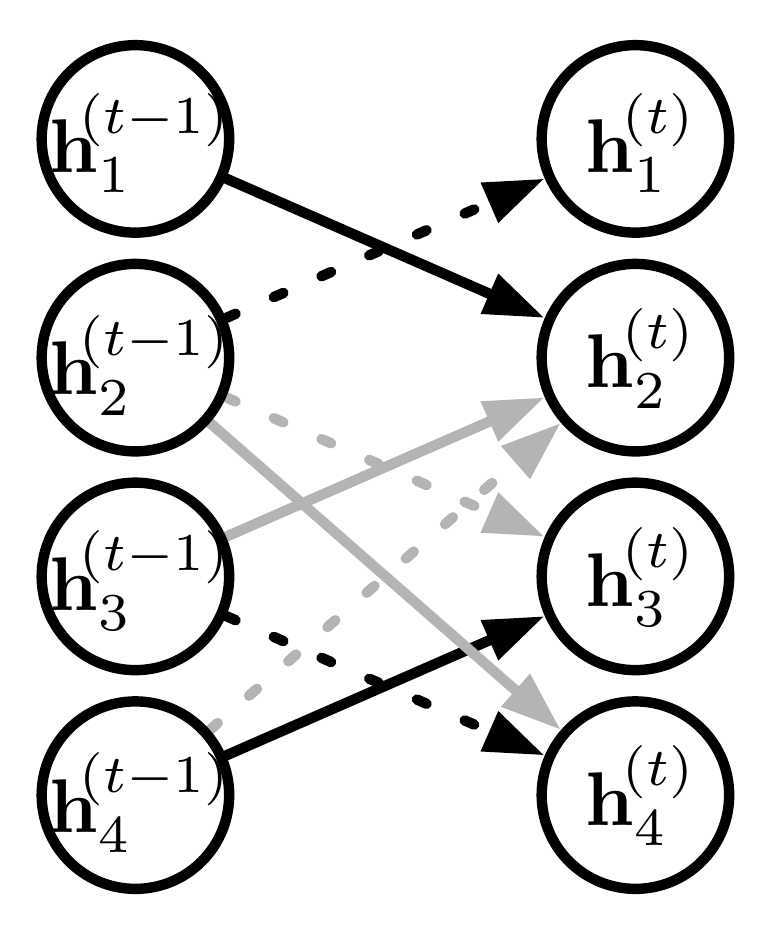} &
\includegraphics[width=.32 \columnwidth]{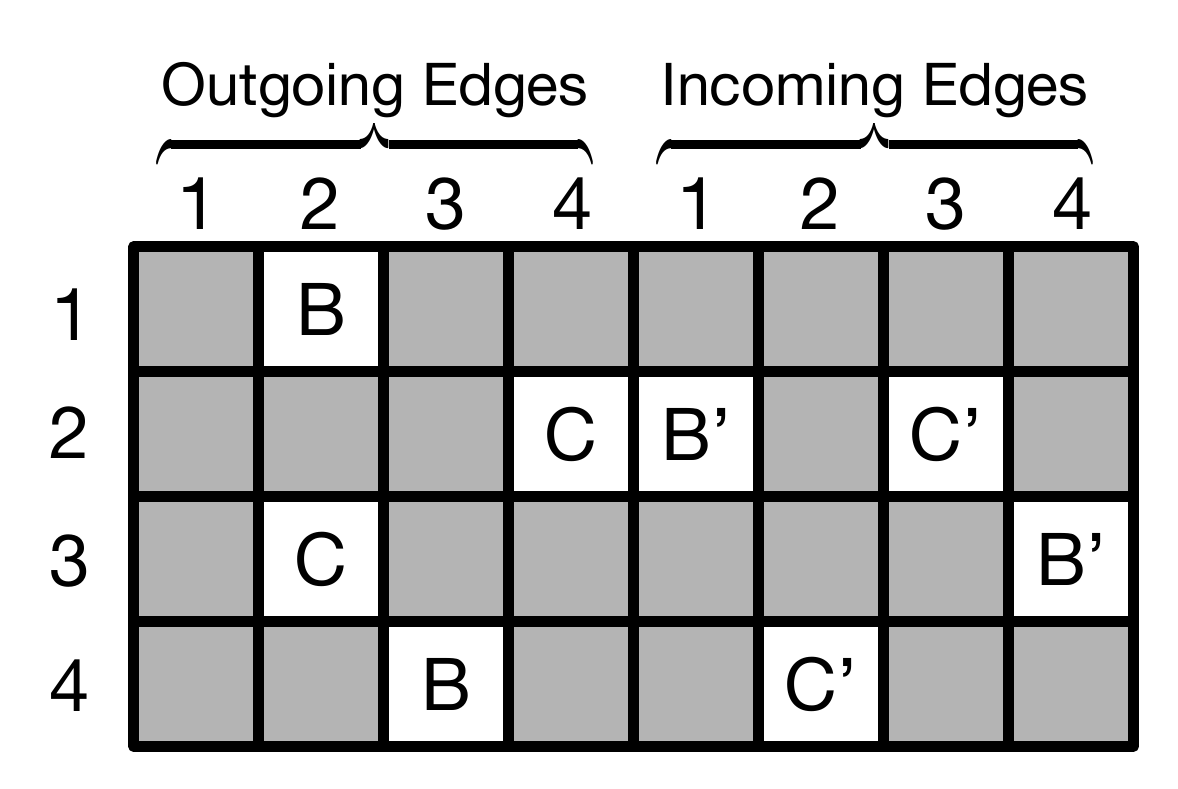}
\\
(a) & (b) & (c) $\Av = \left[ \Av^{(\hbox{\tiny{out}})},  \Av^{(\hbox{\tiny{in}})} \right]$
\end{tabular}
\end{center}
\vspace{-10pt}
\caption{
  (a) Example graph. Color denotes edge types.
  (b) Unrolled one timestep.
  (c) Parameter tying and sparsity in recurrent matrix. Letters denote
  edge types with $B'$ corresponding to the reverse edge of type $B$.
  $B$ and $B'$ denote distinct parameters.
}
\label{fig:graphs-and-sparsity}
\end{figure}

\subsection{Propagation Model}

The basic recurrence of the propagation model is 
\begin{center}
\small
\begin{minipage}{.48\linewidth}
\begin{align}
    \noderept{\node}{1} & = [\labelsv_\node^\top, \mathbf{0}]^\top \label{eq:init}\\
    \nodeactt{\node}{t} & = 
    \Av_{\node:}^\top  \left[\noderept{1}{t-1}{}^\top \ldots
\noderept{|\nodes|}{t-1}{}^\top\right]^\top + \bv \label{eq:propagate-A} \\
    \updategates_{\node}^{t} & =
    \sigma\left(\weights{}^{\updategate}\nodeactt{\node}{t} +
    \selfweights^{\updategate}\noderept{\node}{t-1}\right) \label{eq:update-gate}
\end{align}
\end{minipage}
\hfill
\begin{minipage}{.48\linewidth}
\begin{align}
    \resetgates_{\node}^{t} & =
    \sigma\left(\weights{}^{\resetgate}\nodeactt{\node}{t} +
    \selfweights^{\resetgate}\noderept{\node}{t-1}\right) \label{eq:reset-gate} \\
    \widetilde{\noderept{\node}{t}} & =
    \transform\left(\weights{}\nodeactt{\node}{t} +
    \selfweights\left(\resetgates_{\node}^{t}\odot\noderept{\node}{t-1}\right)\right)
    \\
    \noderept{\node}{t} & = (1 - \updategates_{\node}^{t})\odot
    \noderept{\node}{t-1} + \updategates_{\node}^{t}\odot
    \widetilde{\noderept{\node}{t}}.
\end{align}
\end{minipage}
\end{center}

The matrix $\Av \in \reals^{D|\nodes| \times 2 D |\nodes|}$ determines
how nodes in the graph communicate with each other. The sparsity
structure and parameter tying in $\Av$ is illustrated in
\figref{fig:graphs-and-sparsity}. The sparsity structure corresponds to the
edges of the graph, and the parameters in each
submatrix are determined by the edge type and direction.
$\Av_{\node:} \in \reals^{D|\nodes| \times 2 D}$ are the two columns of blocks
in $\Av^{(\mathrm{out})}$ and $\Av^{(\mathrm{in})}$ corresponding to node $\node$.
\eqref{eq:init} is the initialization step, which copies node annotations
into the first components of the hidden state and pads the rest with
zeros.  \eqref{eq:propagate-A} is the step that passes information
between different nodes of the graph via incoming and outgoing edges
with parameters dependent on the edge type and direction.
$\nodeactt{\node}{t}\in\reals^{2D}$ contains activations from edges in both
directions.
The remaining are GRU-like updates that incorporate information from the
other nodes and from the previous timestep to update each node's
hidden state.  $\updategates$ and $\resetgates$ are the update and
reset gates, $\sigma(x)=1/(1+e^{-x})$ is the logistic sigmoid
function, and $\odot$ is element-wise multiplication. 
We initially experimented with
a vanilla recurrent neural network-style update, but in preliminary
experiments we found this GRU-like propagation step to be more
effective.

\subsection{Output Models}

There are several types of one-step outputs that we would like to
produce in different situations. 
First, \OurMethodMinorShorts{} support \emph{node selection} tasks by making 
$o_{\node} = g(\noderept{\node}{T}, \labelsv_{\node})$ for each node $v
\in \nodes$
output node scores and applying a softmax over node scores.
Second, for graph-level outputs, we define a graph level representation vector as
\begin{align}
  \noderep{\graph} & = \tanh \left(\sum_{\node \in \nodes}
  \sigma\left(i(\noderept{\node}{T}, \labelsv_{\node})\right) \odot
  \tanh \left(j(\noderept{\node}{T}, \labelsv_{\node})\right)\right),
  \label{eq:graph-representation}
\end{align}
where $\sigma(i(\noderept{\node}{T}, \labelsv_{\node}))$ acts as a soft attention
mechanism that decides which nodes are relevant to the current graph-level
task. $i$ and $j$ are 
neural networks that
take the concatenation of $\noderept{\node}{T}$ and $\labelsv_\node$ as input and outputs
real-valued vectors. The $\tanh$ functions can also be replaced with the identity.

\section{\OurMethods}

Here we describe \OurMethods~(\OurMethodShorts), in which several
\OurMethodMinorShorts{} operate in sequence to produce an output sequence
$\outToken{1} \ldots \outToken{K}$.

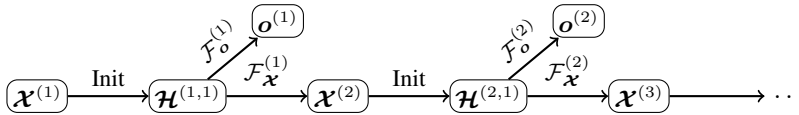
\begin{figure}[t]
  \begin{center}
  \begin{tikzpicture}
    [hstate/.style={draw,rectangle,rounded corners,inner sep=2pt,font=\small},
     updateLabel/.style={sloped,font=\footnotesize}]
    \def\colSep{2.0}
    \def\rowSep{1}
    \node[hstate] (L1)  at (0,0)                         {$\LL{1}$};
    \node[hstate] (H11) at ($(L1)  + (\colSep,0)$)       {$\HH{1}{1}$};
    \node[hstate] (O1)  at ($(H11) + (.6*\colSep,\rowSep)$) {$\outToken{1}$};
    \node[hstate] (L2)  at ($(H11) + (\colSep,0)$)       {$\LL{2}$};
    \node[hstate] (H21) at ($(L2)  + (\colSep,0)$)       {$\HH{2}{1}$};
    \node[hstate] (O2)  at ($(H21) + (.6*\colSep,\rowSep)$) {$\outToken{2}$};
    \node[hstate] (L3)  at ($(H21) + (\colSep,0)$)       {$\LL{3}$};
    \node[]       (cnt) at ($(L3)  + (\colSep,0)$)       {$\ldots$};
    \path[->,thick]
      (L1)  edge node[above,updateLabel] {Init}          (H11)
      (H11) edge node[above,updateLabel] {\GNNOut{1}}    (O1)
      (H11) edge node[above,updateLabel] {\GNNLabel{1}}  (L2)
      (L2)  edge node[above,updateLabel] {Init}          (H21)
      (H21) edge node[above,updateLabel] {\GNNOut{2}}    (O2)
      (H21) edge node[above,updateLabel] {\GNNLabel{2}}  (L3)
      (L3)  edge                                         (cnt)
    ;
  \end{tikzpicture}
  \end{center}
  \vspace{-2ex}
  \caption{Architecture of \OurMethodShort~models.}
  \vspace{-2ex}
  \label{fig:seq-architecture2}
\end{figure}

For the $k^{th}$ output step, we denote the matrix of node annotations as
$\LL{k} = [\labelsv_{1}^{(k)}; \ldots; \labelsv_{|\nodes|}^{(k)}]^\top \in
\reals^{|\nodes| \times \numNodeLabels}$.
We use two \OurMethodMinorShorts~\GNNOut{k} and \GNNLabel{k}: \GNNOut{k} for
predicting $\outToken{k}$ from $\LL{k}$, and \GNNLabel{k} for
predicting $\LL{k+1}$ from $\LL{k}$. $\LL{k+1}$ can be seen as the states
carried over from step $k$ to $k+1$. Both \GNNOut{k} and \GNNLabel{k}
contain a propagation model and an output model. In the propagation models, we denote the matrix of node vectors at the $t^{th}$ propagation step of the
$k^{th}$ output step as
$\HH{k}{t} = [\noderept{1}{k,t}; \ldots; \noderept{|\nodes|}{k,t}]^\top \in \reals^{|\nodes| \times D}$.
As before, in step $k$, we set $\HH{k}{1}$ by $0$-extending $\LL{k}$ per node.
An overview of the model is shown in \figref{fig:seq-architecture2}.
Alternatively, $\GNNOut{k}$ and $\GNNLabel{k}$ can share a single propagation
model, and just have separate output models. This simpler variant is faster to
train and evaluate, and in many cases can achieve similar performance level as
the full model.  But in cases where the desired propagation behavior for
$\GNNOut{k}$ and $\GNNLabel{k}$ are different, this variant may not work as
well.


We introduce a \emph{node annotation} output model for predicting $\LL{k+1}$
from $\HH{k}{T}$. The prediction is done for each node independently using
a neural network $j(\noderept{\node}{k,T}, \labelsv_\node^{(k)})$ that takes the
concatenation of $\noderept{\node}{k,T}$ and $\labelsv_\node^{(k)}$ as input and outputs a vector of
real-valued scores:
\begin{equation}
    \labelsv_\node^{(k+1)} = \sigma\left(j(\noderept{\node}{k,T},
    \labelsv_\node^{(k)})\right).
\end{equation}

There are two settings for training \OurMethodShorts: specifying
all intermediate annotations $\LL{k}$, or training the full model end-to-end given only
$\LL{1}$, graphs and target sequences.
The former can improve performance when we have domain knowledge about specific
intermediate information that should be represented in the internal state of
nodes, while the latter is more general. We describe both.

\paragraph{Sequence outputs with observed annotations}

Consider the task of making a sequence of predictions for a graph, where each
prediction is only about a part of the graph. In order to
ensure we predict an output for each part of the graph exactly once, it
suffices to have one bit per node, indicating whether the node has been
``explained'' so far.
In some settings, a small number of annotations are sufficient
to capture the state of the output procedure. When this is the case, we
may want to directly input this information into the model via labels
indicating target intermediate annotations. In some cases, these annotations
may be \emph{sufficient}, in that we can define a model where
the \OurMethodMinorShorts~are rendered
conditionally independent given the annotations.

In this case, at training time, given the annotations $\LL{k}$ the sequence prediction task
decomposes into single step prediction tasks and can be trained as separate \OurMethodMinorShorts.
At test time, predicted annotations from one step will be used as
input to the next step. This is analogous to training directed graphical models when
data is fully observed.

\paragraph{Sequence outputs with latent annotations}

More generally, when intermediate node annotations $\LL{k}$ are not available during
training, we treat them as hidden units in the network, and train the whole
model jointly by backpropagating through the whole sequence.

\comment{
\begin{figure}
    \begin{center}
        \includegraphics[width=0.8\columnwidth]{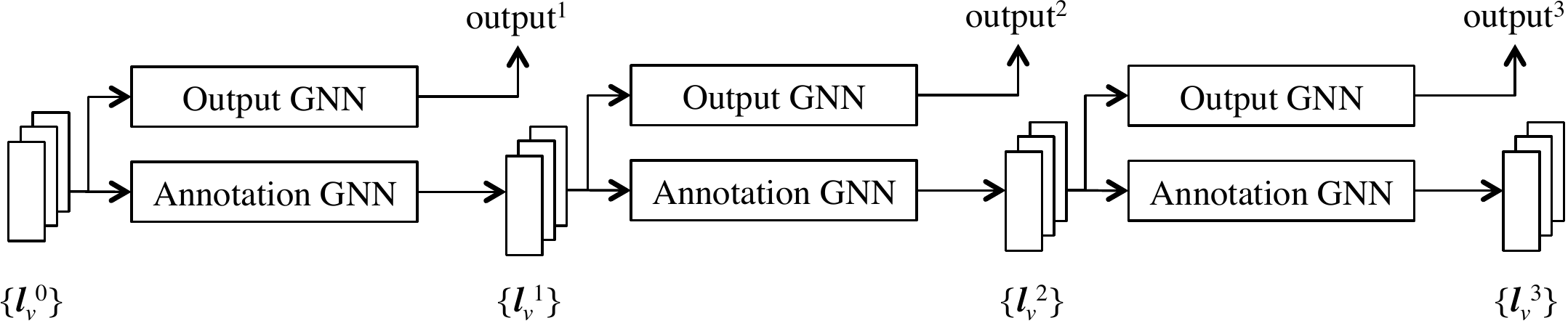}
    \end{center}
    \caption{Architecture of our \OurMethodShorts~model. The initial node
    attributes $\labelsv_\node^0$ are filled with task related information.  The
    node attribute vectors for intermediate steps can be fully observed,
    partially observed, or fully unobserved (this is the case when we use the
    annotation model to predict the full $\labelsv_\node$).
    \TODO{interesting that we don't need to feed output $t-1$ in at timestep
      $t$. Presumably this is because there is little uncertainty over the
      outputs? E.g., if there were two equally likely outputs, how would we
      know which one we started producing?}
  \TODO{Replace GNN with \OurMethodMinorShort}
    }
    \label{fig:seq-architecture}
\end{figure}
}





\section{Explanatory Applications}

In this section we present example applications that concretely illustrate the use of \OurMethodShorts.
We focus on a selection of bAbI artificial intelligence (AI) tasks \citep{weston2015towards} and two graph algorithm learning tasks.

\subsection{bAbI Tasks}
The bAbI tasks are meant to test reasoning capabilities that AI systems should be capable of.
In the bAbI suite, there are 20 tasks that test basic forms of reasoning like deduction, induction, counting,
and path-finding.

We have defined a basic transformation procedure that maps bAbI tasks
to \OurMethodMinorShorts~or \OurMethodShorts. We use the \texttt{--symbolic} option from the
released bAbI code to get stories that just involve sequences
of relations between entities, which are then converted into a graph. Each entity is mapped to a
node, and each relation is mapped to an edge with edge label given by the
relation. The full story is consumed and mapped to a single
graph.  Questions are marked by \texttt{eval} in the data and are comprised
of a question type (e.g., \texttt{has\_fear}), and some
argument (e.g., one or more nodes). The arguments are converted into
initial node annotations, with the $i$-th bit of the $i$-th argument node's
annotation vector set to 1.
For example, if the \texttt{eval} line is \texttt{eval E > A true}, then \texttt{E} gets
initial annotation $\labelsv_E^{(1)}=[1,0]^\top$, \texttt{A} gets
$\labelsv_A^{(1)}=[0,1]^\top$, and for all other nodes $\node$,
$\labelsv_\node^{(1)}=[0,0]^\top$. Question type is 1 (for
`\texttt{>}') and output is class 1 (for `true').
Some tasks have multiple question types, for example Task 4 which has 4
question types: \texttt{e}, \texttt{s}, \texttt{w}, \texttt{n}. For such tasks
we simply train a separate \OurMethodMinorShort~for each task.
We do not use the strong supervision labels or give the \OurMethodShorts~any
intermediate annotations in any experiments.

While simple, this transformation does not preserve all
information about the story (e.g., it discards temporal order of the
inputs), and it does not easily handle ternary and higher order
relations (e.g., \texttt{Yesterday John went to the garden} is not
easily mapped to a simple edge).
We also emphasize that it is a non-trivial task to map general natural language to symbolic form,\footnote{Although the bAbI data is quite templatic, so it is straightforward to hand-code a parser that will work for the bAbI data; the symbolic option removes the need for this.}
so we could not directly apply this approach to arbitrary natural language.
Relaxing these restrictions is left for future work.

However, even with this simple transformation, there are a variety of
bAbI tasks that can be formulated, including Task 19 (Path Finding),
which is arguably the hardest task. We provide baselines to show that
the symbolic representation does not help RNNs or LSTMs significantly,
and show that \OurMethodShorts~solve the problem with a small number
of training instances. We also develop two new bAbI-like tasks that
involve outputting sequences on graphs: shortest paths, and a simple
form of Eulerian circuits (on random connected 2-regular graphs).
The point of these experiments is to illustrate the capabilities
of \OurMethodShorts~across a variety of problems.

\paragraph{Example 1. }
As an example, below is an instance from the symbolic dataset for bAbI task 15, Basic
Deduction.
\begin{framed}
\begin{alltt}
D is A
B is E
A has_fear F
G is F
E has_fear H
F has_fear A
H has_fear A
C is H
eval B has_fear      H
eval G has_fear      A
eval C has_fear      A
eval D has_fear      F
\end{alltt}
\end{framed}
Here the first 8 lines describe the facts, the \OurMethodMinorShort~will use
these facts to build a graph.  Capital letters are nodes, \texttt{is} and
\texttt{has\_fear} are interpreted as edge labels or edge types. The last 4
lines are 4 questions asked for this input data. \texttt{has\_fear} in these
lines are interpreted as a question type. For this task, in each question only
one node is special, e.g. the \texttt{B} in \texttt{eval B has\_fear}, and we
assign a single value 1 to the annotation vector for this special node and 0
to all the other nodes.

For RNN and LSTM the data is converted into token
sequences like below:
\begin{framed}
    \texttt{n6 e1 n1 eol n6 e1 n5 eol n1 e1 n2 eol n4 e1 n5 eol n3 e1 n4 eol
    n3 e1 n5 eol n6 e1 n4 eol q1 n6 n2 ans 1}
\end{framed}
where \texttt{n<id>} are nodes, \texttt{e<id>} are edges, \texttt{q<id>} are
question types, extra tokens \texttt{eol} (end-of-line) and \texttt{ans}
(answer) are added to give the RNN \& LSTM access to the complete information
available in the dataset. The final number is the class label.

\paragraph{Example 2. }
As a second example, below is an instance from the symbolic dataset for bAbI task 19, Path Finding.
\begin{framed}
\begin{alltt}
E s A
B n C
E w F
B w E
eval path B A w,s
\end{alltt}
\end{framed}
Here the first 4 lines describe edges, \texttt{s}, \texttt{n}, \texttt{w},
\texttt{e} (\texttt{e} does not appear in this example) are all different edge types. The last line
is a path question, the answer is a sequence of directions \texttt{w,s}, as
the path going from \texttt{B} to \texttt{A} is to first go west to \texttt{E}
then go south to \texttt{A}. The \texttt{s}, \texttt{n}, \texttt{w},
\texttt{e} in the question lines are treated as output classes.

\paragraph{More Training Details. }
For all tasks in this section, we generate 1000 training examples and 1000 test examples,
50 of the training examples are used for validation.
When evaluating model performance, for all bAbI tasks that contain more than
one questions in one example, the predictions for different questions were
evaluated independently.  As there is randomness
in the dataset generation process, we generated 10 such datasets for each
task, and report the mean and standard deviation of the evaluation
performance across the 10 datasets.

For all explanatory tasks, we start by training different models on only 50
training examples, and gradually increase the number of training examples to
100, 250, 500, and 950 (50 of the training examples are reserved for
validation) until the model's test accuracy reaches 95\% or above, a success by
bAbI standard \cite{weston2015towards}. For each method, we report the minimum number of
training examples it needs to reach 95\% accuracy along with the accuracy it
reaches with that amount of training examples.
In all these cases, we unrolled the propagation process for 5 steps.
For bAbI task 4, 15, 16, 18, 19, we used \OurMethodMinorShort~with the size of node
vectors $\hv^{(t)}_v$ set to $D=4$, $D=5$, $D=6$, $D=3$ and $D=6$ respectively.
For all the GGS-NNs in this section we used the simpler variant in which
$\GNNOut{k}$ and $\GNNLabel{k}$ share a single propagation model. For shortest path and Eulerian circuit tasks, we used $D=20$.
All models are trained long enough with Adam \citep{kingma2014adam}, and the validation set is used to choose
the best model to evaluate and avoid models that are overfitting.

\subsubsection{Single Step Outputs}
We choose four bAbI tasks that are suited to the restrictions
described above and require single step outputs:~4 (Two Argument Relations), 
15 (Basic Deduction), 16 (Basic Induction), and 18 (Size Reasoning).
For Task 4, 15 and 16, a node selection \OurMethodMinorShort~is used. For Task 18
we used a graph-level classification version. All the GGNN networks
contain less than 600 parameters\footnote{For bAbI task 4, we treated `e',
`s', `w', `n' as 4 question types and trained one \OurMethodMinorShort~for
each question type, so strictly speaking for bAbI task 4 our
\OurMethodMinorShort~model has 4
times the number of parameters of a single \OurMethodMinorShort~model.  In our experiments we
used a \OurMethodMinorShort~with 271 parameters for each question type which means 1084
parameters in total.}.

As baselines, we train RNN and LSTM models on the symbolic
data in raw sequence form. The RNNs and LSTMs use 50 dimensional
embeddings and 50 dimensional hidden layers; they predict a single
output at the end of the sequences and the output is treated as a
classification problem, the loss is cross entropy. The RNNs and LSTMs contain
around 5k and 30k parameters, respectively.

\begin{table}[h]
  \centering
    \begin{tabular}{@{}lccc@{}}
        \toprule
        Task    & RNN   & LSTM  & \OurMethodMinorShort \\
        \midrule
        bAbI Task \phantom{1}4 & 97.3$\pm$1.9 (250) & 97.4$\pm$2.0 (250) &
        100.0$\pm$0.0 (50) \\
        bAbI Task 15 & 48.6$\pm$1.9 (950) & 50.3$\pm$1.3 (950) & 100.0$\pm$0.0 (50) \\
        bAbI Task 16 & 33.0$\pm$1.9 (950) & 37.5$\pm$0.9 (950) & 100.0$\pm$0.0 (50) \\
        bAbI Task 18 & 88.9$\pm$0.9 (950) & 88.9$\pm$0.8 (950) & 100.0$\pm$0.0 (50) \\
        \bottomrule
    \end{tabular}
    \caption{Accuracy in percentage of different models for different tasks.
    Number in parentheses is number of training examples required to reach shown
    accuracy.}
  \label{table:single-step-babi}
\end{table}
Test results appear in Table~\ref{table:single-step-babi}. For all tasks
\OurMethodMinorShort~achieves perfect test accuracy using only 50 training
examples, while the RNN/LSTM baselines either use more training examples (Task
4) or fail to solve the tasks (Task 15, 16 and 18).

In Table~\ref{table:task4-breakdown}, we further break down performance of the baselines for task 4 as the amount
of training data varies. While both the RNN and LSTM are able to solve the task
almost perfectly, the \OurMethodMinorShort~reaches 100\% accuracy with much less
data.

\begin{table}[ht]
    \centering
    \begin{tabular}{@{}lccccc@{}}
        \toprule
        \#Training Examples    & 50    & 100   & 250   & 500   & 950 \\
        \midrule
        RNN  & 76.7$\pm$3.8 & 90.2$\pm$4.0 & 97.3$\pm$1.9 & 98.4$\pm$1.3 & 99.7$\pm$0.4 \\
        LSTM & 73.5$\pm$5.2 & 86.4$\pm$3.8 & 97.4$\pm$2.0 & 99.2$\pm$0.8 & 99.6$\pm$0.8 \\
        \bottomrule
    \end{tabular}
    \caption{Performance breakdown of RNN and LSTM on bAbI task 4 as the
    amount of training data changes.}
  \label{table:task4-breakdown}
\end{table}

\subsubsection{Sequential Outputs}

The bAbI Task 19 (Path Finding) is arguably the hardest task among all bAbI
tasks (see
e.g., \citep{sukhbaatar2015end}, which reports an accuracy of less than 20\% for
all methods that do not use the strong supervision). We apply a \OurMethodShort~to
this problem, again on the symbolic form of the data (so results are not comparable
to those in \citep{sukhbaatar2015end}).
An extra `end' class is added to the end of each output sequence;
at test time the network
will keep making predictions until it predicts the `end' class.


The results for this task are given in Table \ref{table:seq-tasks}.
Both RNN and LSTM fail on this task. However, with only 50
training examples, our \OurMethodShorts~achieve much better test
accuracy than RNN and LSTM.


\subsection{Learning Graph Algorithms}

\begin{table}[h]
\small
    \centering
    \begin{tabular}{@{}l|c|c|ccc@{}}
        \toprule
        Task & RNN & LSTM & \multicolumn{3}{c}{\OurMethodShorts} \\
        \midrule
        bAbI Task 19     & 24.7$\pm$2.7 (950) & 28.2$\pm$1.3 (950) & \phantom{1}71.1$\pm$14.7 (50) & 92.5$\pm$5.9 (100) & 99.0$\pm$1.1 (250) \\
        Shortest Path    & \phantom{0}9.7$\pm$1.7 (950) & 10.5$\pm$1.2 (950) & \multicolumn{3}{l}{100.0$\pm$\phantom{0}0.0 (50)} \\
        Eulerian Circuit & \phantom{0}0.3$\pm$0.2 (950) & \phantom{0}0.1$\pm$0.2 (950) & \multicolumn{3}{l}{100.0$\pm$\phantom{0}0.0 (50)} \\
        \bottomrule
    \end{tabular}
    \caption{Accuracy in percentage of different models for different tasks.
    The number in parentheses is number of training examples required to reach that
    level of accuracy.}
    \label{table:seq-tasks}
\end{table}

We further developed two new bAbI-like tasks based on algorithmic problems
on graphs:~Shortest Paths, and Eulerian Circuits. For the first, we generate
random graphs and produce a story that lists all edges in the graphs. 
Questions come from choosing two random nodes $A$ and $B$ and asking for the shortest
path (expressed as a sequence of nodes) that connects the two chosen nodes.
We constrain the data generation to only produce questions where there is a
unique shortest path from $A$ to $B$ of length at least 2. For Eulerian circuits,
we generate a random two-regular connected graph and a separate random distractor
graph. The question gives two nodes $A$ and $B$ to start  the circuit, then
the question is to return the Eulerian circuit (again expressed as a sequence of nodes)
on the given subgraph that starts by going from $A$ to $B$. Results are shown
in the Table \ref{table:seq-tasks}.  RNN and LSTM fail
on both tasks, but \OurMethodShorts~learns to make perfect predictions using
only 50 training examples.




\section{Program Verification with \OurMethodShorts}
\label{sect:ProgramVerification}

Our work on \OurMethodShorts{} is motivated by a practical application in
program verification.
A crucial step in automatic program verification is the inference of
\emph{program invariants}, which
approximate the set of program states
reachable in an execution.
Finding invariants about data structures is an open problem.
As an example, consider the simple \texttt{C} function on the right.

\begin{wrapfigure}[7]{r}{6.1cm}
\small\vspace{-2ex}
\begin{alltt}
node* concat(node* a, node* b) \{
  if (a == NULL) return b;
  node* cur = a;
  while (cur.next != NULL)
    cur = cur->next;
  cur->next = b;
  return a; \hfill\}
\end{alltt}
\end{wrapfigure}
To prove that this program indeed concatenates the two lists \texttt{a} and
\texttt{b} and that all pointer dereferences are valid, we need to
(mathematically) characterize the program's heap in each iteration of the loop.
For this, we use \emph{separation logic}~\citep{OHearn01,Reynolds02}, which uses
\emph{inductive predicates} to describe abstract data structures.
For example, a \underline{l}ist \underline{s}egment is defined as $\SLls(x, y)
\equiv x = y \lor \exists v, n . \SLls(n, y) \ast x \mapsto \{\texttt{val}:v,
\texttt{next}:n\}$, where $x \mapsto \{\texttt{val}:v, \texttt{next}:n\}$ means
that $x$ points to a memory region that contains a structure with \texttt{val}
and \texttt{next} fields whose values are in turn $v$ and $n$.
The $\ast$ connective is a conjunction as $\land$ in Boolean logic, but
additionally requires that its operators refer to ``separate'' parts of the
heap.
Thus, $\SLls(\texttt{cur}, \texttt{NULL})$ implies that $\texttt{cur}$ is either
$\texttt{NULL}$, or that it points to two values $v, n$ on the heap, where $n$
is described by $\SLls$ again.
The formula $ \exists t . \SLls(\texttt{a}, \texttt{cur}) \ast
\SLls(\texttt{cur}, \texttt{NULL}) \ast\SLls(\texttt{b}, t)$ is an
\emph{invariant} of the loop (i.e., it holds when entering the loop, and
after every iteration).
Using it, we can prove that no program run will fail due to dereferencing an
unallocated memory address (this property is called \emph{memory safety}) and
that the function indeed concatenates two lists using a Hoare-style verification
scheme~\citep{Hoare69}.

The hardest part of this process is coming up with formulas that describe
data structures, and this is where we propose to use machine
learning.  Given a program, we run it a few times and
extract the state of memory (represented as a graph; see below) at relevant program locations,
and then predict a separation logic formula.
Static program analysis tools (e.g., \citep{Piskac14}) can check whether a
candidate formula is sufficient to prove the desired properties (e.g., memory
safety).

\subsection{Formalization}

\paragraph{Representing Heap State as a Graph}
As inputs we consider directed, possibly cyclic graphs representing the heap of
a program. These graphs can be automatically constructed from a program's memory
state.
Each graph node $\heapGNode$ corresponds to an address in memory at which a
sequence of pointers $\heapGNode_0, \ldots, \heapGNode_k$ is stored (we ignore
non-pointer values in this work).
Graph edges reflect these pointer values, i.e., $\heapGNode$ has edges 
labeled with $0, \ldots, k$  that point to nodes $\heapGNode_0, \ldots,
\heapGNode_k$, respectively.
A subset of nodes are labeled as corresponding to program variables.

An example input graph is displayed as ``Input'' in \figref{fig:annotations2}.
In it, the node id (i.e., memory address) is displayed in the node.
Edge labels correspond to specific fields in the program, e.g., $0$ in our
example corresponds to the \texttt{next} pointer in our example function from
the previous section. For binary trees there are two more types
of pointers \texttt{left} and \texttt{right} pointing to the left and right
children of a tree node.

\paragraph{Output Representation}
Our aim is to mathematically describe the shape of the heap.
In our model, we restrict ourselves to a syntactically restricted version of
separation logic, in which formulas are of the form
 $\exists x_1, \ldots, x_n . a_1 \ast \ldots \ast a_m$,
where each atomic formula $a_i$ is either $\SLls(x, y)$ (a list from $x$ to $y$),
$\SLtree(x)$ (a binary tree starting in $x$), or $\SLempty(x)$ (no data
structure at $x$).
Existential quantifiers are used to give names to heap nodes which are
needed to describe a shape, but not labeled by a program variable.
For example, to describe a ``panhandle list'' (a list that ends in a
cycle), the first list element on the cycle needs to be named.
In separation logic, this can be expressed as $\exists t . \SLls(x, t)
\ast \SLls(t, t)$.

\paragraph{Data}
We can generate synthetic (labeled) datasets for this problem.
For this, we fix a set of predicates such as $\SLls$ and $\SLtree$ (extensions
could consider doubly-linked list segments, multi-trees, $\ldots$) together with
their inductive definitions.
Then we enumerate separation logic
formulas instantiating our predicates using a given set of program variables.
Finally,
for each formula, we enumerate heap graphs satisfying that
formula. The result is a dataset consisting of pairs of heap graphs and
associated formulas that are used by our learning
procedures.

\subsection{Formulation as \OurMethodShorts}

It is easy to obtain
the node annotations for the intermediate prediction steps from the data
generation process. So we train a variant of
\OurMethodShort~with observed annotations (observed at training time; not test time)
to infer formulas from heap graphs.
Note that it is also possible to use an unobserved
\OurMethodShort~variant and do end-to-end learning. The procedure breaks down
the production of a separation logic formula into a sequence of
steps. We first decide whether to declare existential variables, and if so,
choose which node corresponds to the variable.
Once we have declared existentials, we iterate over all variable names
and produce a separation logic formula describing the data
structure rooted at the node corresponding to the current variable.

\begin{figure}
  \begin{tikzpicture}[heapgraph,scale=.9]
  \def\rowSep{.35}
  \def\colSep{.1}
  \def\nodeStep{1}
  \node[colLabel] (labelStep)  at (0,0)                        {Step};
  \node[colLabel] (labelGraph) at ($(labelStep) + (3.05,0)$)   {Labeled Graph};
  \node[colLabel] (labelOut)   at ($(labelGraph) + (2.35,0)$)  {Out\phantom{p}};

  \node[stepLabel,anchor=north]               (0labelStep) at ($(labelStep.south) + (0, -\rowSep)$) {Input};
  \node[anchor=west,list,label=90:\texttt{b}] (0b)         at ($(0labelStep.east) + (\colSep,0)$)   {$1$};
  \node[list]                                 (0bn)        at ($(0b)              + (\nodeStep,0)$) {$2$};
  \node[list]                                 (0t)         at ($(0bn)             + (\nodeStep,0)$) {$3$};
  \node[list]                                 (0tn)        at ($(0t)              + (\nodeStep,0)$) {$4$};
  \node[anchor=west,outLabel]                 (0labelOut)  at ($(0tn.east)        + (\colSep,0)$)   {};
  \path[->, thick]
    (0b)  edge node[below,heapEdgeLabel] {$0$} (0bn)
    (0bn) edge node[below,heapEdgeLabel] {$0$} (0t)
    (0t)  edge node[below,heapEdgeLabel] {$0$} (0tn)
    (0tn) edge[bend right] node[above,heapEdgeLabel] {$0$} (0t)
  ;

  \node[stepLabel,anchor=north]               (1labelStep) at ($(0labelStep.south)+ (0, -\rowSep)$) {Line 3/$(\dagger)$};
  \node[anchor=west,list,label=90:\texttt{b}] (1b)         at ($(1labelStep.east) + (\colSep,0)$)   {$1$};
  \node[list]                                 (1bn)        at ($(1b)              + (\nodeStep,0)$) {$2$};
  \node[list,label=90:$\phantom{t}$]          (1t)         at ($(1bn)             + (\nodeStep,0)$) {$3$};
  \node[list]                                 (1tn)        at ($(1t)              + (\nodeStep,0)$) {$4$};
  \node[anchor=west,outLabel]                 (1labelOut)  at ($(1tn.east)        + (\colSep,0)$)   {};
  \path[->, thick]
    (1b)  edge node[below,heapEdgeLabel] {$0$} (1bn)
    (1bn) edge node[below,heapEdgeLabel] {$0$} (1t)
    (1t)  edge node[below,heapEdgeLabel] {$0$} (1tn)
    (1tn) edge[bend right] node[above,heapEdgeLabel] {$0$} (1t)
  ;

  \node[stepLabel,anchor=north]               (2labelStep) at ($(1labelStep.south)+ (0, -\rowSep)$) {Line 4-7/$(\ddagger)$};
  \node[anchor=west,list,label=90:\texttt{b}] (2b)         at ($(2labelStep.east) + (\colSep,0)$)   {$1$};
  \node[list]                                 (2bn)        at ($(2b)              + (\nodeStep,0)$) {$2$};
  \node[list,label=90:$t$]                    (2t)         at ($(2bn)             + (\nodeStep,0)$) {$3$};
  \node[list]                                 (2tn)        at ($(2t)              + (\nodeStep,0)$) {$4$};
  \node[anchor=west,outLabel]                 (2labelOut)  at ($(2tn.east)        + (\colSep,0)$)   {$\exists t .$};
  \path[->, thick]
    (2b)  edge node[below,heapEdgeLabel] {$0$} (2bn)
    (2bn) edge node[below,heapEdgeLabel] {$0$} (2t)
    (2t)  edge node[below,heapEdgeLabel] {$0$} (2tn)
    (2tn) edge[bend right] node[above,heapEdgeLabel] {$0$} (2t)
  ;

  \node[stepLabel,anchor=north]               (3labelStep) at ($(2labelStep.south)+ (0, -\rowSep)$) {Line 10 (for \texttt{b})};
  \node[anchor=west,list,active,label=90:\texttt{b}] (3b)  at ($(3labelStep.east) + (\colSep,0)$)   {$1$};
  \node[list]                                 (3bn)        at ($(3b)              + (\nodeStep,0)$) {$2$};
  \node[list,label=90:$t$]                    (3t)         at ($(3bn)             + (\nodeStep,0)$) {$3$};
  \node[list]                                 (3tn)        at ($(3t)              + (\nodeStep,0)$) {$4$};
  \node[anchor=west,outLabel]                 (3labelOut)  at ($(3tn.east)        + (\colSep,0)$)   {$\exists t .$};
  \path[->, thick]
    (3b)  edge node[below,heapEdgeLabel] {$0$} (3bn)
    (3bn) edge node[below,heapEdgeLabel] {$0$} (3t)
    (3t)  edge node[below,heapEdgeLabel] {$0$} (3tn)
    (3tn) edge[bend right] node[above,heapEdgeLabel] {$0$} (3t)
  ;

  \draw[-] ($(labelOut.north east) + (.1, 0)$) -- ($(labelOut.north east) + (.1, -4.6)$);
  \node[colLabel] (labelStep')  at ($(labelOut.east) + (1.6,0)$) {Step};
  \node[colLabel] (labelGraph') at ($(labelStep') + (3.05,0)$)   {Labeled Graph};
  \node[colLabel] (labelOut')   at ($(labelGraph') + (2.6,0)$)  {Out\phantom{p}};

  \node[stepLabel,anchor=west]                (4labelStep) at ($(0labelStep.east) + (4.9, 0)$)      {Line 11/$(\star)$};
  \node[anchor=west,list,active,label=90:\texttt{b}] (4b)  at ($(4labelStep.east) + (\colSep,0)$)   {$1$};
  \node[list]                                 (4bn)        at ($(4b)              + (\nodeStep,0)$) {$2$};
  \node[list,label=90:$t$]                    (4t)         at ($(4bn)             + (\nodeStep,0)$) {$3$};
  \node[list]                                 (4tn)        at ($(4t)              + (\nodeStep,0)$) {$4$};
  \node[anchor=west,outLabel]                 (4labelOut)  at ($(4tn.east)        + (\colSep,0)$)   {$\exists t . \phantom{\SLls(\texttt{b},t)}$};
  \path[->, thick]
    (4b)  edge node[below,heapEdgeLabel] {$0$} (4bn)
    (4bn) edge node[below,heapEdgeLabel] {$0$} (4t)
    (4t)  edge node[below,heapEdgeLabel] {$0$} (4tn)
    (4tn) edge[bend right] node[above,heapEdgeLabel] {$0$} (4t)
  ;

  \node[stepLabel,anchor=west]                (5labelStep) at ($(1labelStep.east) + (4.9, 0)$)      {Line 13,14/$(\heartsuit)$};
  \node[anchor=west,list,active,label=90:\texttt{b}] (5b)  at ($(5labelStep.east) + (\colSep,0)$)   {$1$};
  \node[list]                                 (5bn)        at ($(5b)              + (\nodeStep,0)$) {$2$};
  \node[list,label=90:$t$]                    (5t)         at ($(5bn)             + (\nodeStep,0)$) {$3$};
  \node[list]                                 (5tn)        at ($(5t)              + (\nodeStep,0)$) {$4$};
  \node[anchor=west,align=left,outLabel]      (5labelOut)  at ($(5tn.east)        + (\colSep,0)$)   {$\exists t . \SLls(\texttt{b},t) \ast$};
  \path[->, thick]
    (5b)  edge node[below,heapEdgeLabel] {$0$} (5bn)
    (5bn) edge node[below,heapEdgeLabel] {$0$} (5t)
    (5t)  edge node[below,heapEdgeLabel] {$0$} (5tn)
    (5tn) edge[bend right] node[above,heapEdgeLabel] {$0$} (5t)
  ;

  \node[stepLabel,anchor=west]                (6labelStep) at ($(2labelStep.east) + (4.9, 0)$)      {Line 18/$(\spadesuit)$};
  \node[anchor=west,list,explained,label=90:\texttt{b}] (6b) at ($(6labelStep.east)+(\colSep,0)$)   {$1$};
  \node[list,explained]                       (6bn)        at ($(6b)              + (\nodeStep,0)$) {$2$};
  \node[list,label=90:$t$]                    (6t)         at ($(6bn)             + (\nodeStep,0)$) {$3$};
  \node[list]                                 (6tn)        at ($(6t)              + (\nodeStep,0)$) {$4$};
  \node[anchor=west,align=left,outLabel]      (6labelOut)  at ($(6tn.east)        + (\colSep,0)$)   {$\exists t . \SLls(\texttt{b},t) \ast$};
  \path[->, thick]
    (6b)  edge node[below,heapEdgeLabel] {$0$} (6bn)
    (6bn) edge node[below,heapEdgeLabel] {$0$} (6t)
    (6t)  edge node[below,heapEdgeLabel] {$0$} (6tn)
    (6tn) edge[bend right] node[above,heapEdgeLabel] {$0$} (6t)
  ;

  \node[stepLabel,anchor=west]                (7labelStep) at ($(3labelStep.east) + (4.9, 0)$)      {Line 10 (for $t$)};
  \node[anchor=west,list,explained,label=90:\texttt{b}] (7b) at ($(7labelStep.east)+(\colSep,0)$)   {$1$};
  \node[list,explained]                       (7bn)        at ($(7b)              + (\nodeStep,0)$) {$2$};
  \node[list,active,label=90:$t$]             (7t)         at ($(7bn)             + (\nodeStep,0)$) {$3$};
  \node[list]                                 (7tn)        at ($(7t)              + (\nodeStep,0)$) {$4$};
  \node[anchor=west,align=left,outLabel]      (7labelOut)  at ($(7tn.east)        + (\colSep,0)$)   {$\exists t . \SLls(\texttt{b},t) \ast$};
  \path[->, thick]
    (7b)  edge node[below,heapEdgeLabel] {$0$} (7bn)
    (7bn) edge node[below,heapEdgeLabel] {$0$} (7t)
    (7t)  edge node[below,heapEdgeLabel] {$0$} (7tn)
    (7tn) edge[bend right] node[above,heapEdgeLabel] {$0$} (7t)
  ;
\end{tikzpicture}
 \vspace{-4ex}
 \caption{Illustration of the first 8 steps to predict a separation logic formula from
   a memory state. Label \emph{is-named} signified by variable near node,
   \emph{active} by double border, \emph{is-explained} by white fill.}
 \label{fig:annotations2}
\end{figure}
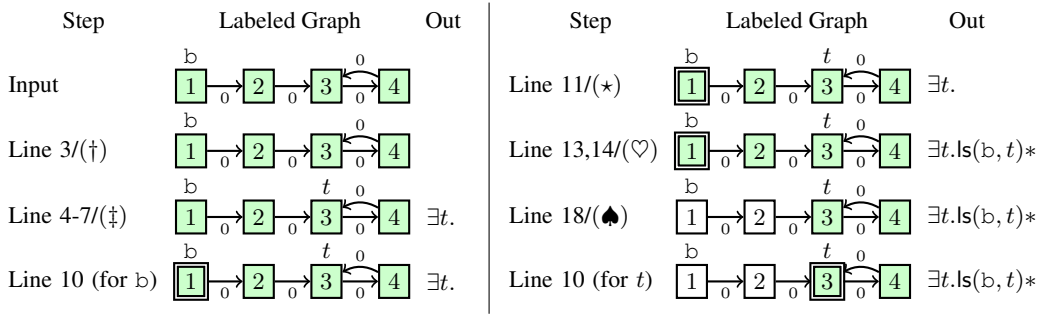

The full algorithm for predicting separation logic formula appears below, as
\algoref{alg:seplogic-prediction}.
We use three explicit node annotations, namely \emph{is-named} (heap node labeled by
program variable or declared existentially quantified variable), \emph{active} (cf. algorithm) and \emph{is-explained} (heap node is
part of data structure already predicted).
Initial node labels can be directly computed from the input graph: ``is-named''
is on for nodes labeled by program variables, ``active'' and ``is-explained''
are always off (done in line 2).
The commented lines in the algorithm are implemented using a
\OurMethodMinorShort, i.e., \algoref{alg:seplogic-prediction} is an instance of
our \OurMethodShort{} model.
An illustration of the beginning of a run of the algorithm is shown in
\figref{fig:annotations2}, where each step is related to one line of the
algorithm.

\begin{algorithm}
  \begin{algorithmic}[1]
    \Require Heap graph $\graph$ with named program variables
    \State{$\LLSym{} \leftarrow $ compute initial labels from $\graph$}
    \State{$\HHSym{} \leftarrow $ initialize node vectors by $0$-extending
      $\LLSym{}$}
    \While{$\exists$ quantifier needed}\Comment{\textbf{Graph-level Classification ($\dagger$)}}
      \State{$t \leftarrow$ fresh variable name}
      \State{$v \leftarrow$ pick node}
        \Comment{\textbf{Node Selection ($\ddagger$)}}      
      \State{$\LLSym{} \leftarrow$ turn on ``is-named'' for $v$ in $\LLSym{}$}
      \State{\algoutput{} ``$\exists t . $''}
    \EndWhile
    \For{node $v_\ell$ with label ``is-named'' in $\LLSym{}$}
      \State{$\HHSym{} \leftarrow$ initialize node vectors, turn on ``active'' label for $v_\ell$ in $\LLSym{}$}
      \State{$\mathit{pred} \leftarrow$ pick data structure predicate}
        \Comment{\textbf{Graph-level Classification ($\star$)}}
      \If{$\mathit{pred} = \SLls$}
        \State{$\ell_{\mathit{end}} \leftarrow$ pick list end node}
          \Comment{\textbf{Node Selection ($\heartsuit$)}}
        \State{\algoutput{} ``$\SLls(\ell, \ell_{\mathit{end}})~\ast$''}
      \Else{}
        \State{\algoutput{} ``$\mathit{pred}(\ell)~\ast$''}
      \EndIf
      \State{$\LLSym{} \leftarrow$ update node annotations in $\LLSym{}$}
        \Comment{\textbf{Node Annotation ($\spadesuit$)}}
    \EndFor
  \end{algorithmic}
  \caption{Separation logic formula prediction procedure}
  \label{alg:seplogic-prediction}
\end{algorithm}


\subsection{Model Setup Details}
We use the full \OurMethodShort~model
where $\GNNOut{k}$ and $\GNNLabel{k}$ have separate propagation models.
For all the \OurMethodMinorShort~components in the \OurMethodShort~pipeline,
we unrolled the propagation process for 10 time steps. The
\OurMethodShorts~associated with step ($\dagger$) (deciding wheter
more existentially quantified variable need to be declared) and ($\ddagger$)
(identify which node need to be declared as existentially quantified) uses
$D=16$ dimensional node representations.  For all other
\OurMethodShort~components, $D=8$ is used.  Adam \citep{kingma2014adam} is
used for optimization, the models are trained on minibatches of 20 graphs, and
optimized until training error is very low. For the graph-level classification tasks, we also
artificially balanced classes to have even number of examples from each class
in each minibatch. All the \OurMethodShort~components contain less than 5k
parameters and no overfitting is observed during training.

\subsection{Batch Prediction Details}
\label{appendix:batch-prediction}

In practice, a set of heap graphs will be given as input and a single output
formula is expected to describe and be consistent with all the input graphs.
The different heap graphs can be snapshots of the heap state at different
points in the program execution process, or different runs of the same program
with different inputs. We call this the ``batch prediction'' setup contrasting
with the single graph prediction described in the main paper.

To make batch predictions, we run one \OurMethodShort~for each graph
simultaneously. For each prediction step, the outputs of all the \OurMethodShorts~at
that step across the batch of graphs are aggregated.

For node selection outputs, the common named
variables link nodes on different graphs togeter, which is the key for
aggregating predictions in a batch.  We compute the score for a particular named
variable $t$ as $o_t = \sum_g o^g_{\nodes_g(t)}$, where $\nodes_g(t)$ maps
variable name $t$ to a node in graph $g$, and $o^g_{\nodes_g(t)}$ is the output score for named
variable $t$ in graph $g$. When applying a softmax over all names using $o_t$
as scores, this is equivalent to a model that computes $p(toselect =
t)=\prod_g p_g(toselect=\nodes_g(t))$.

For graph-level classification outputs, we add up scores of a particular class
across the batch of graphs, or equivalently compute $p(class=k)=\prod_g
p_g(class=k)$.
Node annotation outputs are updated for each graph independently as different
graphs have completely different set of nodes.  However, when the algorithm tries to
update the annotation for one named variable, the nodes
associated with that variable in all graphs are updated.
During training, all labels for intermediate steps are available to us from the data
generation process, so the training process again can be decomposed to single
output single graph training.


A more complex scenario allowing for nested data structures (e.g., list of lists) was discussed in
\cite{brockschmidt2015learning}.
We have also successfully extended the \OurMethodShort~model to this case.
More details on this can be found in Appendix
\ref{appendix:nested-prediction}.

\subsection{Experiments. }
\label{sec:program-verification-results}

For this paper, we produced a dataset of 327 formulas that involves three
program variables, with 498 graphs per formula, yielding around 160,000
formula/heap graph combinations.
To evaluate, we split the data into training, validation and test
sets using a 6:2:2 split on the formulas (i.e., the formulas in the test set
were not in the training set). We measure correctness by
whether the formula predicted at test time is logically equivalent to the
ground truth; equivalence is approximated by canonicalizing names and order
of the formulas and then comparing for exact equality.

We compared our \OurMethodShort-based model with a method we developed
earlier~\citep{brockschmidt2015learning}.
The earlier approach treats each prediction step as standard classification,
and requires complex, manual, problem-specific feature
engineering, to achieve
an accuracy of 89.11\%.
In contrast, our new model was trained with no feature engineering and very
little domain knowledge and achieved an accuracy of 89.96\%.

\begin{figure}
    \begin{center}
        \includegraphics[width=0.5\textwidth]{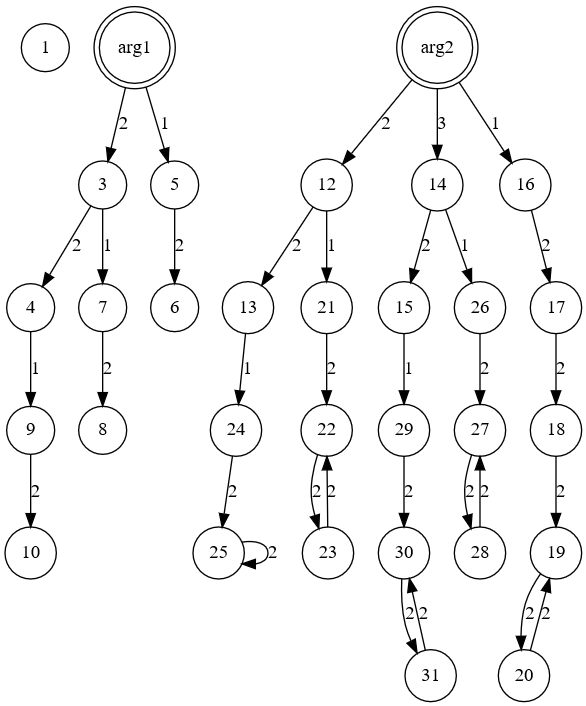}
    \end{center}
    \caption{A heap graph example that contains two named variables
        \texttt{arg1} and \texttt{arg2}, and one isolated \SLnull~node (node 1). All the
    edges to \SLnull~are not shown here for clarity. The numbers on edges
    indicate different edge types.  Our \OurMethodShort~model successfully
    finds the right
formula $\SLls(\texttt{arg1}, \SLnull, \lambda t_1 \rightarrow \SLls(t_1,
\SLnull, \SLNone)) * \SLtree(\texttt{arg2}, \lambda t_2 \rightarrow \exists
e_1. \SLls(t_2, e_1, \SLNone) * \SLls(e_1, e_1, \SLNone))$.}
    \label{fig:appendix-heap-graph-example}
\end{figure}

An example heap graph and the corresponding separation logic formula found by
our \OurMethodShort~model is shown in
\figref{fig:appendix-heap-graph-example}. This example also involves nested
data structures and the batching extension developed in the previous section.

We have also successfully used our new model in a program verification
framework, supplying needed program invariants to a theorem prover to prove
correctness of a collection of list-manipulating algorithms such as insertion
sort.
The following Table \ref{table:seplogic-list-invariants} lists a set of benchmark list manipulation programs and
the separation logic formula invariants found by the \OurMethodShort~model,
which were successfully used in a verification framework to prove the
correctness of corresponding programs. A further extension of the current
pipeline has been shown to be able to successfully prove more sophisticated
programs like sorting programs and various other list-manipulating programs.

\begin{table}
\begin{center}
\begin{tabular}{ll}
\toprule
Program & Invariant Found \\
\midrule
Traverse1 & $\SLls(\SLlst,\SLcurr) * \SLls(\SLcurr,\SLnull)$ \\
Traverse2 & $ \SLcurr \ne \SLnull *
\SLlst \ne \SLnull * \SLls(\SLlst,\SLcurr) * \SLls(\SLcurr,\SLnull)$ \\
Concat & $\texttt{a} \ne \SLnull * a \ne b * b \ne \SLcurr * \SLcurr\ne
\SLnull$ \\
& $* \SLls(\SLcurr, \SLnull) * \SLls(a, \SLcurr) * \SLls(b, \SLnull)$ \\
Copy & $\SLls(\SLcurr, \SLnull) * \SLls(\SLlst, \SLcurr) * \SLls(\texttt{cp},
\SLnull) $ \\
Dispose & $\SLls(\SLlst, \SLnull)$ \\
Insert & $\SLcurr\ne\SLnull * \SLcurr\ne \SLelt * \SLelt\ne \SLnull * \SLelt\ne
\SLlst* \SLlst\ne \SLnull$ \\
& $* \SLls(\SLelt, \SLnull) * \SLls(\SLlst, \SLcurr) * \SLls(\SLcurr, \SLnull)$ \\
Remove & $\SLcurr\ne \SLnull * \SLlst\ne \SLnull * \SLls(\SLlst, \SLcurr) *
\SLls(\SLcurr, \SLnull)$ \\
\bottomrule
\end{tabular}
\caption{Example list manipulation programs and the separation logic formula
invariants the \OurMethodShort~model founds from a set of input graphs. The
``$\ne$'' parts are produced by
a deterministic procedure that goes through all the named program variables in
all graphs and checks for inequality.}
\label{table:seplogic-list-invariants}
\end{center}
\end{table}


\section{Related Work}

The most closely related work is GNNs, which we have discussed at
length above.
\cite{micheli2009neural} proposed another
closely related model that differs from GNNs mainly in the output model.
GNNs have been applied in several domains
\citep{gori2005new,di2006comparison,scarselli2009graph,uwents2011neural}, but they
do not appear to be in widespread use in the ICLR community.
Part of our aim here is to publicize GNNs as a useful and interesting
neural network variant. 

An analogy can be drawn between 
our adaptation from GNNs to \OurMethodMinorShorts, to the work of
\cite{domke2011parameter} and \cite{stoyanov2011empirical} in the structured prediction setting.
There belief propagation (which must be run to near convergence to
get good gradients) is replaced with truncated belief propagation updates, and
then the model is trained so that the truncated iteration produce good
results after a fixed number of iterations.
Similarly, Recursive Neural Networks
\citep{goller1996learning,socher2011parsing} being extended to Tree LSTMs
\citep{tai2015improved} is analogous to our using of GRU updates in
\OurMethodMinorShorts~instead of the standard GNN recurrence with the aim of
improving the long-term propagation of information across a graph structure.

The general idea expressed in this paper of assembling
problem-specific neural networks as a composition of learned
components has a long history, dating back at least to the work of
\cite{hinton1988representing} on assembling neural networks according
to a family tree structure in order to predict relations between
people. Similar ideas appear in \cite{hammer2004neural} and \cite{bottou2014machine}.

Graph kernels
\citep{shervashidze2011weisfeiler,kashima2003marginalized} can be used
for a variety of kernel-based learning tasks with graph-structured
inputs, but we are not aware of work that learns the kernels and
outputs sequences.  \cite{perozzi2014deepwalk} convert graphs into sequences by
following random walks on the graph then learns node embeddings using
sequence-based methods.
\cite{sperduti1997supervised} map graphs to graph vectors then
classify using an output neural network.
There are several models that make use of similar propagation of node
representations on a graph structure.
\cite{bruna2013spectral} generalize
convolutions to graph structures.  The difference between their work
and GNNs is analogous to the difference between convolutional and
recurrent networks.
\cite{duvenaud2015convolutional} also consider convolutional like
operations on graphs, building a learnable, differentiable variant of
a successful graph feature.
%
\cite{lusci2013deep} converts an arbitrary
undirected graph to a number of different DAGs with different
orientations and then propagates node representations inwards towards
each root, training an ensemble of models.
In all of the above, the focus is on one-step problems.

GNNs and our extensions have many of the same desirable properties of
pointer networks \citep{vinyals2015pointer};  when using
node selection output layers, nodes from the input can be chosen as outputs.
There are two main differences:~first,
in GNNs the graph structure is explicit, which makes the models less general
but may provide stronger generalization ability; 
second, pointer networks require that each node has
properties (e.g., a location in space), while GNNs can represent
nodes that are defined only by their position in the graph, which 
makes them more general along a different dimension.

\OurMethodShorts~are related to soft alignment and attentional models
(e.g., \cite{bahdanau2014neural,kumar2015ask,sukhbaatar2015end}) in two respects:~first, the graph
representation in \eqref{eq:graph-representation} uses context to
focus attention on which nodes are important to the current decision;
second, node annotations in the program verification example keep
track of which nodes have been explained so far, which gives an
explicit mechanism for making sure that each node in the input
has been used over the sequence of producing an output.


\section{Discussion}
\label{sec:discussion}

\paragraph{What is being learned?}

\begin{wrapfigure}[4]{r}{3cm}
\small\vspace{-4ex}
\begin{alltt}
 B is E
 E has_fear H
 eval B has_fear
\end{alltt}
\end{wrapfigure}
It is instructive to consider what is being learned by the
\OurMethodMinorShorts. To do so, we can draw analogy between how the bAbI task 15
would be solved via
a logical formulation. As an example, consider the subset of lines
needed to answer one example on the right.

 
To do logical reasoning, we would need not only a logical encoding of
the facts present in the story but also the background world knowledge encoded as
inference rules such as 
\begin{align}
\texttt{is(x, y)} \wedge \texttt{has-fear(y, z)} \implies \texttt{has-fear(x, z)}.
\end{align}
Our encoding of the tasks simplifies the parsing of the story into
graph form, but it does not provide any of the background knowledge. 
The \OurMethodMinorShort~model can be seen as learning this,
with results stored in the neural network weights. 

\comment{
To do logical reasoning, we would need not only a logical encoding of
the facts present in the story but also the background world knowledge encoded as
inference rules such as 
\begin{align}
\texttt{is(x, y)} \wedge \texttt{has-fear(y, z)} \implies \texttt{has-fear(x, z)}.
\end{align}
Our encoding of the tasks simplifies the parsing of the story into
graph form, but it does not provide any of the background world knowledge.
Thus, the \OurMethodMinorShort~model can be seen as learning the background world knowledge,
with results stored in the neural network weights.
}

\paragraph{Discussion}

The results in the paper show that \OurMethodShorts~have
desirable inductive biases across a range of problems that
have some intrinsic graph structure to them, and we believe
there to be many more cases where \OurMethodShorts~will be
useful. There are, however, some limitations that need to be overcome to
make them apply even more broadly. Two limitations that we mentioned
previously are that the bAbI task translation does not incorporate
temporal order of inputs or ternary and higher order relations.
We can imagine several possibilities for lifting these restrictions, such
as concatenating a series of \OurMethodMinorShorts, where there is one
\OurMethodMinorShorts~for each edge, and representing
higher order relations as factor graphs. 
A more significant challenge is how to handle less structured input
representations. For example, in the bAbI tasks it would be desirable
not to use the symbolic form of the inputs. One possible approach
is to incorporate less structured inputs, and latent vectors, in our 
\OurMethodShorts.
However, experimentation is
needed to find the best way of addressing these issues.


The current \OurMethodShorts~formulation specifies a question only after
all the facts have been consumed. This implies that the network must try
to derive all consequences of the seen facts and store all pertinent information
to a node within its node representation. This is likely not ideal;
it would be preferable to develop methods that take the question as an initial
input, and then dynamically derive the facts needed to answer the question.

We are optimistic about the further applications of \OurMethodShorts.
We are particularly interested in continuing to develop end-to-end learnable
systems that can learn about semantic properties of programs, that can
learn more complicated graph algorithms,  and in
applying these ideas to problems that
require reasoning over knowledge bases and databases.
More generally, we consider these graph neural networks as representing a
step towards a model that can combine structured representations with the
powerful algorithms of deep learning, with the aim of taking advantage of known
structure while learning and inferring how to reason with and extend 
these representations.

\section*{Acknowledgements}
We thank Siddharth Krishna, Alex Gaunt,
Emine Yilmaz, Milad Shokouhi,
and Pushmeet Kohli for useful
conversations and Douwe Kiela for comments on an earlier draft of the
paper.

\bibliography{refs}
\bibliographystyle{iclr2016_conference}


\newpage
\appendix

\section{Contraction Map Example}
\label{appendix:contraction-example}

Consider a linear 1-hidden unit cycle-structured GNN with $N$ nodes
$\{1,\dots, N\}$. For simplicity we ignored all edge labels and node labels,
equivalently this is a simple example with $\numNodeLabels=1$ and
$\numEdgeLabels=1$.
At each timestep $t$ we update hidden states $h_1, \ldots, h_N$
as
\begin{align}
  h_i^{(t)} & = m_{i} \cdot h_{i-1}^{(t-1)} + b_{i}, \label{eq:simple-recurrence}
\end{align}
for each $i$, where $m_i$ and $b_i$ are parameters of the propagation model.  We use the convention that
$h_{j}$ cycles around and refers to $h_{N+j}$
when $j \le 0$.
Let $\hv^{(t)} = [h^{(t)}_1, \ldots, h^{(t)}_N]^\top$,
\begin{align}
  \Mv = \left[
    \begin{array}{ccccc}
      0   & 0      & 0      & \ldots  & m_1 \\
      m_2 & 0      & 0      &         & 0 \\
      0   & m_3    & 0      &         & 0 \\
          & \vdots &        & \ddots  &  \\
      0   & 0      &        & m_{N}   & 0 
    \end{array}
    \right]
\end{align}
and $\bv = [b_1, \ldots, b_{N}]^\top$.
We can write the joint update for all $i$ as
\begin{align}
  \hv^{(t)} = \Mv \hv^{(t-1)} + \bv = T(\hv^{(t-1)})
\end{align}
Restrict the update to define a contraction mapping in the
Euclidean metric. This means that there is some $\rho < 1$ such that for any $\hv, \hv'$,
\begin{align}\label{eq:contraction-map-definition}
||T(\hv) - T(\hv') || < \rho ||\hv - \hv'||,
\end{align}
or in other words,
\begin{align}
|| \Mv (\hv - \hv') || < \rho ||\hv - \hv'||.
\end{align}
We can immediately see that this implies that $|m_i| < \rho$ for
each $i$ by letting $\hv$ be the elementary vector that is all
zero except for a 1 in position $i-1$ and letting $\hv'$ be the all zeros vector.

Expanding Eq.~\ref{eq:simple-recurrence}, we get
\begin{align}
  h_i^{(t)} & = m_{i} \cdot (m_{i-1} h_{i-1}^{(t-2)} + b_{i-1}) + b_{i}
    \nonumber\\
            & = m_{i} m_{i-1} h_{i-1}^{(t-2)} + m_{i} b_{i-1} + b_{i}
    \nonumber\\
            & = m_{i} m_{i-1} (m_{i-2} h_{i-2}^{(t-3)} + b_{i-2}) + m_{i}
    b_{i-1} + b_{i} \nonumber\\
            & = m_{i} m_{i-1} m_{i-2} h_{i-2}^{(t-3)} + m_{i} m_{i-1} b_{i-2} + m_{i} b_{i-1} + b_{i}.
\end{align}
In the GNN model, node label $l_i$ controls which values of $m_i$
and $b_i$ are used during the propagation. Looking at this expansion
and noting that $m_i < \rho$ for all $i$, we see that information about
labels of nodes $\delta$ away will decay at a rate of
$\left(\frac{1}{\rho}\right)^\delta$.
Thus, at least in this simple case, the restriction that $T$ be
a contraction means that it is not able to maintain long-range
dependencies.

\subsection{Nonlinear Case}

The same analysis can be applied to a nonlinear update, i.e.
\begin{equation}
    h^{(t)}_i = \sigma\left(m_i \cdot h^{(t-1)}_{i-1} + b_i\right),
\end{equation}
where $\sigma$ is any nonlinear function. Then $T(\hv) =
\sigma\left(\Mv\hv + \bv\right)$. Let $T(\hv)=[T_1(\hv), ...,
T_N(\hv)]^\top$, where $T_i(\hv^{(t-1)}) = h_i^{(t)}$. The contraction map
definition Eq.~\ref{eq:contraction-map-definition} implies that each entry of
the Jacobian matrix of $T$ is bounded by $\rho$, i.e.
\begin{equation}
    \left|\pdiff{T_i}{h_j}\right| < \rho, \qquad \forall i, \forall j.
\end{equation}
To see this, consider two vectors $\hv$ and $\hv'$, where $h_k=h_k', \forall
k\neq j$ and $h_j + \Delta = h_j'$. The definition in
\eqref{eq:contraction-map-definition} implies that for all $i$,
\begin{equation}
    ||T_i(\hv) - T_i(\hv')|| \le ||T(\hv) - T(\hv')|| < \rho |\Delta|.
\end{equation}
Therefore
\begin{equation}
\left\|\frac{T_i(h_1, ..., h_{j-1}, h_j, h_{j+1}, ..., h_N) - T_i(h_1, ...,
h_{j-1}, h_j+\Delta, h_{j+1}, ..., h_N)}{\Delta}\right\| < \rho,
\end{equation}
where the left hand side is $\left\|\pdiff{T_i}{h_j}\right\|$ by definition as
$\Delta\rightarrow 0$.

When $j=i-1$, 
\begin{equation}
    \left|\pdiff{T_i}{h_{i-1}}\right| < \rho.
\end{equation}
Also, because of the special cycle graph structure, for all other $j$s we have
$\pdiff{T_i}{h_j}=0$.  Applying this to the update at timestep $t$, we get
\begin{equation}
    \left|\pdiff{h^{(t)}_i}{h^{(t-1)}_{i-1}}\right| < \rho.
\end{equation}

Now let's see how a change in $h^{(1)}_1$ could affect $h^{(t)}_t$.  Using
the chain rule and the special graph structure, we have
\begin{align}
    \left|\pdiff{h^{(t)}_t}{h^{(1)}_1}\right| &=
    \left|\pdiff{h^{(t)}_t}{h^{(t-1)}_{t-1}}\cdot
    \pdiff{h^{(t-1)}_{t-1}}{h^{(t-2)}_{t-2}}\cdots
    \pdiff{h^{(2)}_2}{h^{(1)}_1}\right| \nonumber\\
    &= 
    \left|\pdiff{h^{(t)}_t}{h^{(t-1)}_{t-1}}\right|\cdot
    \left|\pdiff{h^{(t-1)}_{t-1}}{h^{(t-2)}_{t-2}}\right|\cdots
    \left|\pdiff{h^{(2)}_2}{h^{(1)}_1}\right| \nonumber\\
    &< \rho\cdot \rho\cdots \rho = \rho^{t-1}.
\end{align}
As $\rho < 1$, this derivative will approach 0 exponentially fast as $t$
grows. Intuitively, this means that the impact one node has on another node
far away will decay exponetially, therefore making it difficult to model long
range dependencies.

\section{Why are RNN and LSTM so Bad on the Sequence Prediction Tasks?}

RNN and LSTM performance on the sequence prediction tasks, i.e. bAbI task 19,
shortest path and Eulerian circuit, are very poor compared to single output
tasks.  The Eulerian circuit task is the one that RNN and LSTM fail most
dramatically.  A typical training example for this task looks like the
following,
\begin{framed}
\begin{alltt}
3 connected-to 7
7 connected-to 3
1 connected-to 2
2 connected-to 1
5 connected-to 7
7 connected-to 5
0 connected-to 4
4 connected-to 0
1 connected-to 0
0 connected-to 1
8 connected-to 6
6 connected-to 8
3 connected-to 6
6 connected-to 3
5 connected-to 8
8 connected-to 5
4 connected-to 2
2 connected-to 4
eval eulerian-circuit 5 7       5,7,3,6,8
\end{alltt}
\end{framed}
This describes a graph with two cycles 3-7-5-8-6 and 1-2-4-0, where 3-7-5-8-6
is the target cycle and 1-2-4-0 is a smaller distractor graph.  All edges are
presented twice in both directions for symmetry. The task is to find the cycle
that starts with the given two nodes and in the direction from the first to
the second.  The distractor graph is added to increase the difficulty of this
task, this also makes the output cycle not strictly ``Eulerian''.

For RNN and LSTM the above training example is further transformed into a
sequence of tokens,
\begin{framed}
\begin{alltt}
n4 e1 n8 eol n8 e1 n4 eol n2 e1 n3 eol n3 e1 n2 eol n6 e1 n8 eol
n8 e1 n6 eol n1 e1 n5 eol n5 e1 n1 eol n2 e1 n1 eol n1 e1 n2 eol
n9 e1 n7 eol n7 e1 n9 eol n4 e1 n7 eol n7 e1 n4 eol n6 e1 n9 eol
n9 e1 n6 eol n5 e1 n3 eol n3 e1 n5 eol q1 n6 n8 ans 6 8 4 7 9
\end{alltt}
\end{framed}
Note the node IDs here are different from the ones in the original symbolic
data. The RNN and LSTM read through the whole sequence, and start to predict
the first output when reading the \texttt{ans} token.  Then for each
prediction step, the \texttt{ans} token is fed as the input and the target
node ID (treated as a class label) is expected as the output.  In this
current setup, the output of each prediction step is not fed as the input for
the next. Our \OurMethodShort~model uses the same setup, where the output
of one step is not used as input to the next, only the predicted node
annotations $\LL{k}$ carry over from one step to the next, so the
comparison is still fair for RNN and LSTM.  Changing both our method and the
baselines to make use of previous predictions is left as future work.

From this example we can see that the sequences the RNN and LSTM have to
handle is quite long, close to 80 tokens before the predictions are made. Some
predictions really depend on long range memory, for example the first edge
(3-7) and first a few tokens (\texttt{n4 e1 n8}) in the sequence are needed to
make prediction in the third prediction step (3 in the original symbolic data,
and 4 in the tokenized RNN data). Keeping long range memory in RNNs is
challenging, LSTMs do better than RNNs but still can't completely solve the
problem.

Another challenge about this task is the output sequence does not appear in
the same order as in the input sequence.  In fact, the data has no sequential
nature at all, even when the edges are randomly permutated, the target output
sequence should not change.  The same applies for bAbI task 19 and the shortest
path task. \OurMethodShorts~are good at handling this type of ``static'' data,
while RNN and LSTM are not.  However future work is needed to determine how
best to apply \OurMethodShorts~to temporal sequential data which RNN and LSTM are good at.
This is one limitation of the \OurMethodShorts~model which we discussed in
\secref{sec:discussion}.

\section{Nested Prediction Details}
\label{appendix:nested-prediction}

Data structures like \emph{list of lists} are nested data structures, in which
the \texttt{val} pointer of each node in a data structure points to another data
structure.  Such data structures can
be represented in separation logic by allowing predicates to be nested.  For
example, a list of lists can be represented as $\SLls(x, y, \lambda
t\rightarrow\SLls(t, \texttt{NULL}))$, where $\lambda t\rightarrow\SLls(t,
\texttt{NULL})$ is a lambda expression and says that for each node in the
list from $x$ to $y$, its \texttt{val} pointer $t$ satisfies $\SLls(t,
\texttt{NULL})$. So there is
a list from $x$ to $y$, where each node in that list points to another list. A
simple list without nested structures can be represented as $\SLls(x, y,
\SLNone)$ where $\SLNone$ represents an empty predicate. Note
that unlike the non-nested case where the \texttt{val} pointer always points
to \texttt{NULL}, we have to
consider the \texttt{val} pointers here in order to describe and handle nested data
structures.

To make our \OurMethodShorts~able to predict nested formulas, we adapt
\algoref{alg:seplogic-prediction} to \algoref{alg:seplogic-nesting}. Where
an outer loop goes through each named variable once and generate a nested predicate
with the node associated with that variable as the active node. The nested
prediction procedure handles prediction similarly as in
\algoref{alg:seplogic-prediction}. Before calling the nested prediction
procedure recursively, the node annotation update in line
\ref{alg:line-node-annotation} not only annotates nodes in the current structure
as ``is-explained'', but also annotates nodes linked to via the ``val'' pointer
from all nodes in the current structure as ``active''. For the list of lists
example, after predicting ``$\SLls(x, y,$'', the annotation step annotates all
nodes in the list from $x$ to $y$ as ``is-explained'' and all nodes the
\texttt{val} pointer points to from the list as ``active''.  This knowledge is
not hard coded into the algorithm, the annotation model can learn this
behavior from data.

\begin{algorithm}
  \begin{algorithmic}[1]
    \Procedure{OuterLoop}{$\graph$}
        \Comment{Graph $\graph$ with named program variables}
        \State{$\LLSym{} \leftarrow $ compute initial labels from $\graph$}
        \For{each variable name $\SLvar$}
          \State{$v_\ell\leftarrow$ node associated with $\SLvar$}
          \State{turn on ``active'' bit for $v_\ell$ in $\LLSym{}$}
          \State{\Call{PredictNestedFormula}{$\graph$, $\LLSym{}$, $\SLvar$}}
        \EndFor
    \EndProcedure
    \\
    \Procedure{PredictNestedFormula}{$\graph$, $\LLSym{}$, $\SLvar$}
      \State{$\HHSym{} \leftarrow $ initialize node vectors by $0$-extending
        $\LLSym{}$}
      \While{$\exists$ quantifier needed}\Comment{\textbf{Graph-level Classification ($\dagger$)}}
        \State{$e \leftarrow$ fresh existentially quantified variable name}
        \State{$v \leftarrow$ pick node}
          \Comment{\textbf{Node Selection ($\ddagger$)}}      
        \State{$\LLSym{} \leftarrow$ turn on ``is-named'' for $v$ in $\LLSym{}$}
        \State{\algoutput{} ``$\exists e . $''}
      \EndWhile
      \If{$\SLvar$ is a lambda variable name}
        \State{\algoutput{} ``$\lambda~\SLvar.$''}
      \EndIf
      \State{$\mathit{pred} \leftarrow$ pick data structure predicate}
        \Comment{\textbf{Graph-level Classification ($\star$)}}
      \If{$\mathit{pred} = \SLls$}
        \State{$\ell_{\mathit{end}} \leftarrow$ pick list end node}
          \Comment{\textbf{Node Selection ($\heartsuit$)}}
        \State{$\SLvar_\mathit{end} \leftarrow$ get variable name associated
        with $\ell_\mathit{end}$}
        \State{\algoutput{} ``$\SLls(\SLvar, \SLvar_{\mathit{end}}, $''}
      \ElsIf{$\mathit{pred}=\SLtree$}
        \State{\algoutput{} ``$\SLtree(\SLvar,$''}
      \Else
        \State{\algoutput{} ``$\SLempty(\SLvar)~\ast$''}
        \State{\Return}
      \EndIf
      \State{$\LLSym{} \leftarrow$ update node annotations in $\LLSym{}$}
        \Comment{\textbf{Node Annotation ($\spadesuit$)}}
        \label{alg:line-node-annotation}
      \State{$t \leftarrow$ fresh lambda variable name}
      \State{\Call{PredictNestedFormula}{$\graph$, $\LLSym{}$, $t$}}
        \Comment{Recursively predict all nested formulas.}
      \State{\algoutput{} ``$)~\ast$''}
    \EndProcedure
  \end{algorithmic}
  \caption{Nested separation logic formula prediction procedure}
  \label{alg:seplogic-nesting}
\end{algorithm}


\end{document}